\documentclass[runningheads]{llncs}
\usepackage[T1]{fontenc}
\usepackage{graphicx,verbatim}
\usepackage{amsmath}
\usepackage{amssymb}
\usepackage{booktabs}
\usepackage{multirow}
\usepackage{xcolor}
\newcommand{\methodname}{SutureFormer} %global definiton of method name
\usepackage[colorlinks=true,linkcolor=blue,citecolor=blue,urlcolor=black]{hyperref}
\begin{document}
\title{\methodname: Learning Surgical Trajectories via Goal-conditioned Offline RL in Pixel Space}
\titlerunning{\methodname: Learning Surgical Trajectories}
\author{Huanrong Liu\inst{1,2}$^{,*}$ \and
Chunlin Tian\inst{1}$^{,*}$ \and
Tongyu Jia\inst{3}$^{,*}$ \and
Tailai Zhou\inst{3}$^{,*}$ \and \\
Qin Liu\inst{1} \and
Yu Gao\inst{3} \and
Yutong Ban\inst{7} \and
Yun Gu\inst{5,6} \and \\
Guy Rosman\inst{4}$^{,\dagger}$ \and
Xin Ma\inst{3}$^{,\dagger}$ \and
Qingbiao Li\inst{1,2}$^{,\dagger}$
}
\index{Liu, Huanrong}
\index{Tian, Chunlin}
\index{Jia, Tongyu}
\index{Zhou, Tailai}
\index{Liu, Qin}
\index{Gao, Yu}
\index{Ban, Yutong}
\index{Gu, Yun}
\index{Rosman, Guy}
\index{Ma, Xin}
\index{Li, Qingbiao}
\authorrunning{H. Liu et al.}
\institute{Faculty of Science and Technology, University of Macau, Macau, China \and
University of Macau Advanced Research Institute in Hengqin, Zhuhai, China \and
Senior Department of Urology, The Chinese PLA General Hospital, Beijing, China \and
School of Medicine, Duke University, Durham, North Carolina, USA \and 
Shanghai Key Laboratory of Flexible Medical Robotics, Tongren Hospital, Institute of Medical Robotics, Shanghai Jiao Tong University, Shanghai, China \and
School of Automation and Intelligent Sensing, Shanghai Jiao Tong University, Shanghai, China \and
Global College, Shanghai Jiao Tong University, Shanghai, China \\
\email{qingbiaoli@um.edu.mo}\\
$^{*}$\,Equal contribution.\quad$^{\dagger}$\,Corresponding author.}

\maketitle              % typeset the header of the contribution
%

% Sections
\begin{abstract}

Predicting surgical needle trajectories from endoscopic video is critical for robot-assisted suturing, enabling anticipatory planning, real-time guidance, and safer motion execution. However, most surgical recordings lack synchronized robot kinematics, and dense frame-level annotations are prohibitively expensive to obtain from clinical experts. To address these challenges, we propose \methodname, a goal-conditioned offline reinforcement learning framework that reformulates trajectory prediction as visual navigation in pixel space. \methodname\ operates solely on raw image sequences under sparse supervision, eliminating the need for kinematic signals or dense annotations. It encodes variable-length video observations using a Spatial CNN and Transformer architecture to capture both local spatial cues and long-range temporal dependencies. It then autoregressively predicts future waypoints within an action space comprising discrete directions and continuous magnitudes. A guidance channel constructed from nine sparse keyframe waypoints provides goal conditioning. To enable stable offline training from expert demonstrations, we adopt Conservative Q-Learning with confidence-weighted rewards and behavioral cloning regularization.
We evaluate \methodname\ on a new kidney wound suturing dataset comprising 1,158 trajectories from 50 patients, where it reduces Average Displacement Error by 56.8\% compared to the strongest baseline,  demonstrating the effectiveness of pixel-level sequential action modeling and the offline RL formulation for surgical trajectory prediction. The code will be available in https://github.com/HaroldHuanrongLIU/MICCAI2026\_SutureFormer.


\end{abstract}
\section{Introduction}
\label{sec:introduction}

Robot-assisted surgery is evolving from tele-operation toward task-level autonomy, where intelligent systems anticipate surgical intent and provide proactive assistance~\cite{yang2017medical,attanasio2021autonomy}. 
Within this paradigm, the capacity to predict instrument trajectories, particularly during suturing, one of the most technically demanding and outcome-sensitive maneuvers in minimally invasive procedures. 

From the perspective of bounded rationality~\cite{simon1972theories}, surgeons operate under inherent cognitive constraints: limited attention span, finite working memory, and time pressure collectively bound the optimality of intraoperative decisions. A learning-based trajectory prediction system that distills expert demonstrations into anticipatory guidance can therefore serve as a decision-support mechanism, helping less experienced surgeons approximate expert-level performance and ultimately improving access to high-quality surgical care.

Despite growing interest in deep learning for surgical scene analysis~\cite{maier2017surgical}, including workflow recognition~\cite{jin2022trans} and scene understanding~\cite{nwoye2022rendezvous}, research on precise procedural assistance for trajectory prediction remains nascent. Current methods face two critical practical barriers. First, most approaches depend on robot kinematic signals, including joint angles, end-effector poses, and gripper states~\cite{qin2020davincinet,shi2022recognition,weerasinghe2024multimodal}. Such data are available only on platforms with accessible kinematic interfaces (e.g., the da Vinci Research Kit) and often require direct collaboration with device manufacturers. This dependency fundamentally limits transferability: a policy trained on one robotic platform cannot generalize to another, let alone to the vast archive of conventional laparoscopic video where no kinematic readout exists. In addition, many methods require dense temporal annotations, which are prohibitively expensive to obtain.

Learning control from pixels is a challenging but transformative idea, one that has seen remarkable success in adjacent fields. For instance, purely vision-based imitation learning agents have mastered complex control tasks like driving, learning to navigate and plan by directly mapping image inputs to steering commands~\cite{pomerleau1989alvinn,bojarski2016end}. Cai et al.~\cite{cai2021vision} demonstrated that cost functions for reinforcement learning (RL) can be learned directly from images, bypassing the need for explicit state estimation. Similarly, Tamar et al. proposed Value Iteration Networks (VIN)~\cite{tamar2016value} that a neural network can learn to perform goal-directed reasoning, generalizing its policy to novel environments by embedding a computational process akin to planning within its architecture. These insights suggest that the "kinematic bottleneck" in surgical trajectory prediction is not a technological inevitability, but a design choice.

To address these limitations, we propose \methodname, which reformulates surgical trajectory prediction as goal-conditioned visual navigation in pixel space using offline expert demonstrations. A goal-conditioned policy iteratively predicts future needle waypoints based solely on local visual context and sparse keyframe guidance.
The observation encoder combines a Spatial CNN and Transformer to capture spatial cues and long-range temporal dependencies. A discrete-action Conservative Q-Learning (CQL) agent then autoregressively generates trajectory waypoints using a 9 direction action space with continuous step magnitudes, conditioned on keyframe goals during training and polynomial extrapolation at inference. In practice, \methodname\ requires only endoscopic video frames and 9 sparse keyframes annotations per trajectory. To our knowledge, this is the first framework to formulate surgical trajectory prediction as visual navigation task with offline RL in pixel space. Our method outperforms current state-of-the-art methods, demonstrating the viability of offline RL for surgical trajectory prediction.
\section{Methods}
\label{sec:methods}

\begin{figure}[t]
    \centering
    \includegraphics[width=0.98\linewidth]{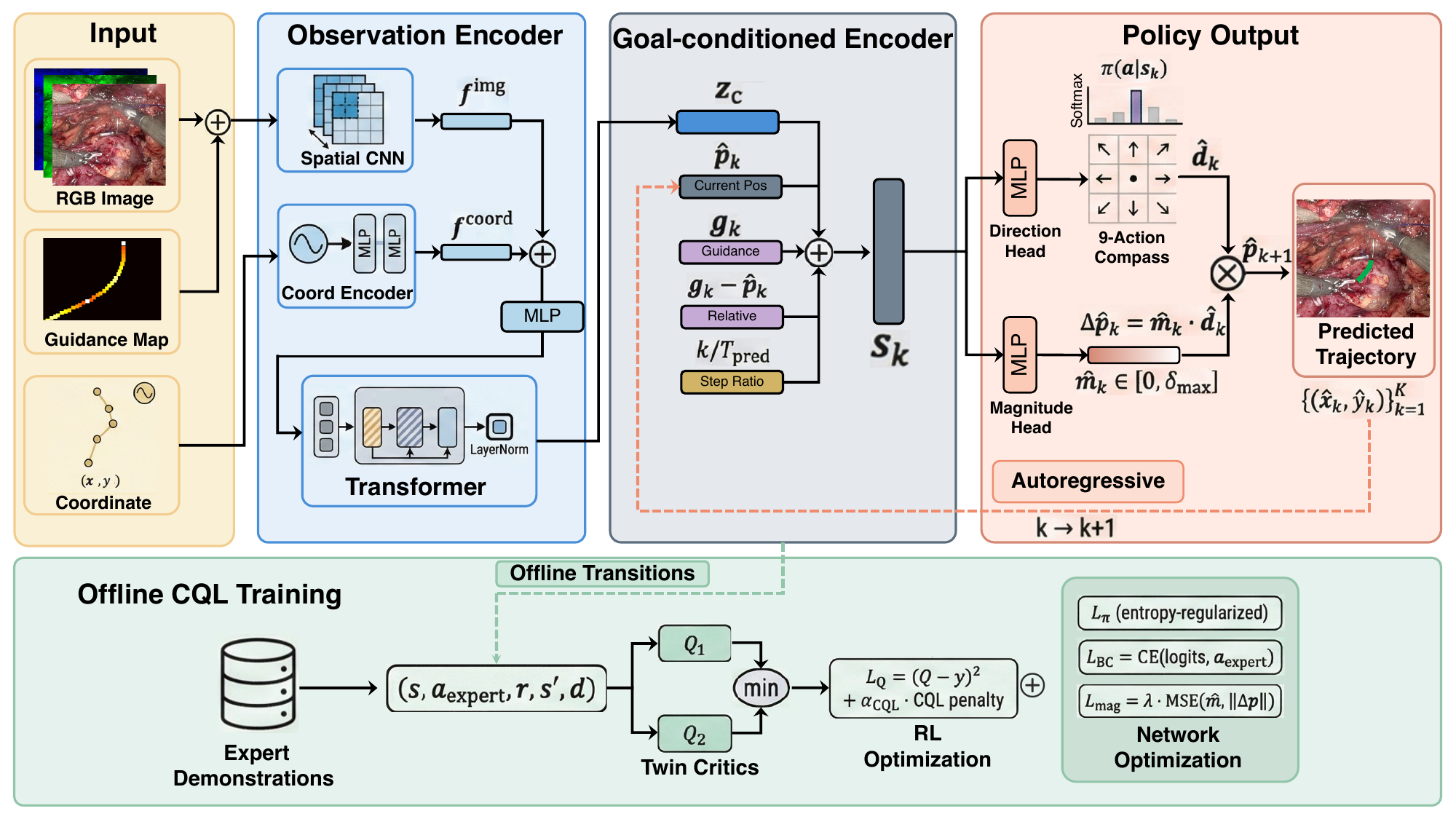}
    \caption{Overview of the proposed framework. The Observation Encoder fuses visual and coordinate features and aggregates them via a Transformer to produce $\mathbf{z}_{\mathrm{c}}$. The Goal-conditioned Encoder combines $\mathbf{z}_{\mathrm{c}}$ with the current position, guidance target, relative displacement, and step ratio into a unified state $\mathbf{s}_k$. The Policy Output predicts direction and magnitude to autoregressively generate the trajectory. The model is trained entirely offline via Conservative Q-Learning with twin critics and supplementary behavior cloning and magnitude supervision losses.}
    \label{fig:architecture}
\end{figure}

\textbf{Problem Statement.} 
Given a variable-length observed video sequence $V_{0:T_P} = \{I_t\}_{t=0}^{T_P}$ with corresponding needle tip coordinates $\{(x_t, y_t)\}_{t=0}^{T_P}$, our goal is to predict the remaining trajectory $\{(\hat{x}_t, \hat{y}_t)\}_{t=T_P+1}^{T_P+T_F}$, where $T_P$ and $T_F$ denote the number of observed and predicted frames, respectively. We formulate this task as a goal-conditioned Markov Decision Process and solve it using offline CQL. Fig.~\ref{fig:architecture} illustrates the overall architecture, which employs an encoder-decoder framework where the decoder functions as a pixel-space reinforcement learning trajectory generator conditioned on the past frames.

\subsection{Observation Encoder}
\label{subsec:observation_encoder}

Each frame is represented by a $128\times 128$ local crop extracted from the full image and centered at the needle tip. The crop is augmented with a guidance channel, a single-channel heatmap encoding the trajectory path confidence at each pixel, and concatenated with the RGB data to form a 4-channel input. The observation encoder comprises: (i) a  SpatialCNN extracts features from all observed input frames $\{I_t\}_{t=0}^{T_P}$ ; (ii) a sinusoidal coordinate encoder that maps all 2D coordinate $\{(x_t, y_t)\}_{t=0}^{T_P}$, normalized to the range $[0, 1]$, then subsequently transformed by a two-layer MLP; and (iii) a Transformer that aggregates the variable-length observation sequence.

Specifically, for each trajectory, an image feature $\mathbf{f}^{\mathrm{img}}$ and a coordinate feature $\mathbf{f}^{\mathrm{coord}}$ are extracted from historical observation, concatenated, and projected into the Transformer dimension of $d_{\mathrm{model}}$. The Transformer employs a causal attention mask and a padding mask to support variable observation lengths $T_{\mathrm{obs}}$. Finally, LayerNorm is applied to the last valid output token to yield the contextual representation $\mathbf{z}_{\mathrm{c}}$.

\subsection{Goal-conditioned Encoder}
\label{sec:goal_navigation}

At each prediction step $k$, a state representation is constructed by combining the observation context with the current navigation state:
\begin{equation}
    \mathbf{s}_k = \phi\big(\mathbf{z}_\mathrm{c},\; \hat{\mathbf{p}}_{k},\; \mathbf{g}_k,\; \mathbf{g}_k - \hat{\mathbf{p}}_{k},\; k / T_\mathrm{pred}\big)
    \label{eq:state}
\end{equation}
where $\mathbf{g}_k$ denotes the guidance coordinate at step $k$. Three independent sinusoidal coordinate encoders are utilized to map the current position $\hat{\mathbf{p}}_{k}$, the guidance target $\mathbf{g}_k$, and the relative displacement $\mathbf{g}_k - \hat{\mathbf{p}}_{k}$. Additionally, a linear layer maps the scalar step ratio $k/T_\mathrm{pred}$ to a feature space. The concatenation $\phi$ of these encoded features with the contextual representation $\mathbf{z}_\mathrm{c}$ yields the final state vector $\mathbf{s}_k$.

The proposed method employs a discrete 9-direction action space. A direction head outputs the logits over these 9 actions, whereas a magnitude head predicts a continuous scalar step magnitude $\hat{m}_k\in[0,\delta_{\max}]$. During inference, the predicted displacement is formulated as:
\begin{equation}
    \hat{\mathbf{d}}_k = \sum_{a=1}^{9}\pi(a|\mathbf{s}_k)\,\mathbf{u}_a
    \Delta\hat{\mathbf{p}}_k = \hat{m}_k\,\hat{\mathbf{d}}_k, \qquad
    \hat{\mathbf{p}}_{k+1}=\mathrm{clip}(\hat{\mathbf{p}}_k+\Delta\hat{\mathbf{p}}_k,0,1)
    \label{eq:displacemnet}
\end{equation}
where $\hat{\mathbf{d}}_k = \sum_{a=1}^{9}\pi(a|\mathbf{s}_k)\,\mathbf{u}_a$ represents the softmax-weighted direction vector that produces smooth motion, and $\mathbf{u}_a$ denotes the unit vector corresponding to action $a$. 

During training, the guidance coordinates $\mathbf{g}_k$ are constructed by placing confidence-weighted annotated trajectory points within the $128 \times 128$ crop, applying dilation via a $3 \times 3$ max filter, and concatenating the result with the RGB data to form a 4-channel input. During testing, due to the unavailability of future trajectory points, pseudo-guidance is generated via polynomial extrapolation from the observed points. All the positions are clamped to the range of $[0,1]^2$ to avoid exceeding the normalized image boundary.

\subsection{Offline Conservative Q-Learning}
\label{subsec:offline_cql}

\methodname\ is trained entirely offline using a static dataset of expert demonstrations. During the training phase, the ground-truth trajectories are traversed to extract expert transitions. At each step $k$, the spatial displacement $\mathbf{p}_{k+1} - \mathbf{p}_{k}$ is discretized into the nearest compass direction via cosine similarity (assigned an idle action if $\|\mathbf{p}_{k+1} - \mathbf{p}_{k}\| < 10^{-6}$), and the corresponding expert magnitude is defined as $m_k^* = \|\mathbf{p}_{k+1} - \mathbf{p}_{k}\|_2$.

The proposed framework adopts Conservative Q-Learning (CQL)~\cite{kumar2020cql}, which incorporates a penalty into the standard Bellman backup to mitigate the overestimation of the Q-function for out-of-distribution actions. By penalizing the Q-values associated with unobserved actions, this approach ensures that the learned policy remains proximate to the demonstrated expert behavior. The critic loss for a given transition $(\mathbf{s},a_{\mathrm{expert}},r,\mathbf{s}',d)$ is formulated as
\begin{equation}
    \mathcal{L}_Q = (Q(\mathbf{s}, a_{\mathrm{expert}}) - y)^2 + \alpha_{\mathrm{CQL}} \cdot \left(\log \sum_{a'} \exp Q(\mathbf{s}, a') - Q(\mathbf{s}, a_{\mathrm{expert}})\right)
    \label{eq:cql}
\end{equation}
where $y = r + \gamma^n (1 - d) \cdot V(\mathbf{s}')$ represents the $n$-step target with a discount factor $\gamma = 0.95$, whereas the parameter $\alpha_{\mathrm{CQL}} = 0.01$ controls the strength of the conservative regularization. The maximum-entropy soft value target $V(\mathbf{s}')$ is computed as:
\begin{equation}
    V(\mathbf{s}')=\sum_{a'}\pi(a'|\mathbf{s}')
    \left[\min\!\left(Q_1^{\mathrm{tgt}}(\mathbf{s}',a'),Q_2^{\mathrm{tgt}}(\mathbf{s}',a')\right)
    -\alpha\log\pi(a'|\mathbf{s}')\right]
    \label{eq:value}
\end{equation}
where $\alpha=0.2$ denotes a fixed entropy temperature. Furthermore, the actor is optimized via an entropy-regularized objective function utilizing the estimated Q-values:
\begin{equation}
    \mathcal{L}_\pi = \mathbb{E}_{\mathbf{s}}\left[\sum_{a} \pi(a \mid \mathbf{s}) \left(\alpha \log \pi(a \mid \mathbf{s}) - \min(Q_1(\mathbf{s}, a),\, Q_2(\mathbf{s}, a))\right)\right]
    \label{eq:policy}
\end{equation}
This objective is supplemented by a behavior cloning loss $\mathcal{L}_{\mathrm{BC}} = \mathrm{CE}(\mathrm{logits}, a_{\mathrm{expert}})$ with a weighting coefficient $\lambda_{\mathrm{BC}} = 1.0$, and a supervised magnitude loss $\mathcal{L}_{\mathrm{mag}} = \lambda_{\mathrm{mag}} \cdot \mathrm{MSE}(\hat{m}, \|\mathbf{p}_{\mathrm{target}} - \mathbf{p}_{\mathrm{current}}\|_2)$ with $\lambda_{\mathrm{mag}} = 100$.

We note a critical design choice of separating gradient flows. Specifically, the Q-networks receive states with detached gradients to prevent the propagation of gradients into the observation encoder. Conversely, the losses associated with the actor and the step magnitude allow gradients to propagate through the state encoder and into the contextual representation $\mathbf{z}_{\mathrm{c}}$. This configuration ensures that the objectives of trajectory prediction, rather than those of value estimation, dictate the formation of the visual representations.

\subsection{Reward Design with Sparse Keyframe Supervision}
\label{subsec:reward}

At each prediction step $k$, the reward comprises three components: a constant time penalty $r_{\mathrm{time}}=-0.01$ encouraging efficient trajectories, a confidence-weighted proximity reward based on the distance $d_k$ to the ground truth, and an exponential terminal bonus at the final step. The proximity reward is positive (up to $r_{\mathrm{prox,max}} = 0.5$) when $d_k$ falls within a threshold of 0.02 in normalized coordinates, and negative otherwise. Since clinical experts annotate only 9 keyframes per trajectory while intermediate positions are obtained via temporal interpolation, we apply confidence-based weighting: keyframe steps receive full weight ($w = 1.0$), whereas interpolated steps are down-weighted proportionally ($w = 0.5 + 0.5 \cdot confidence$, see detail in Sec.\ref{subsec:interpolated annotation}), providing dense training signals while reflecting the lower certainty of non-keyframe positions. To handle variable-length observation and prediction sequences, we employ masked attention within the Transformer encoder to selectively attend to valid time steps while ignoring padded positions. Furthermore, we use bucketed batch sampling, which groups sequences of similar lengths into the same batch, thereby minimizing padding overhead and improving training efficiency.
\section{Experiments}
\label{sec:experiments}

\noindent\textbf{Datasets.}
We evaluate \methodname\ on real clinical dataset of robotic-assisted laparoscopic kidney wound suturing operated by expert surgeon. The dataset comprises surgical videos from 50 patients and contains a total of 1{,}158 trajectories, where clinical experts annotate 9 keyframes within each trajectory. The dataset is partitioned at the patient level into training, validation, and test sets, consisting of 35, 8, and 7 patients (corresponding to 861, 151, and 146 trajectories), respectively.

\noindent\textbf{Interpolated Annotation.}
\label{subsec:interpolated annotation}
To obtain dense per-frame annotations, we independently apply cubic spline interpolation~\cite{de1978practical} to the $x$ and $y$ coordinate sequences as functions of the frame index. Given the 9 keyframe positions, we fit two natural cubic spline functions, $S_x(t)$ and $S_y(t)$, evaluating them at every intermediate frame within the temporal range. The interpolated coordinates are rounded to the nearest integer to yield valid pixel locations, without extrapolation beyond the first and last keyframes. Each interpolated frame receives a confidence score from 0.45 to 0.9 based on its temporal proximity to the nearest keyframe, assigning higher scores to closer frames. The expert annotated keyframes retain a confidence score of 1.0.

\noindent\textbf{Implementation Details.}
The model is trained for 100 epochs with batch size 8 on an NVIDIA RTX 5090 GPU. We use four Adam optimizers with cosine annealing decaying to 1\% of the initial learning rates: $1 \times 10^{-4}$ for the observation encoder and $3\times 10^{-4}$ for the actor, critic, and magnitude heads. CQL hyperparameters are $\alpha_\mathrm{CQL} = 0.01$, $\gamma = 0.95$, and $n=3$-step returns, with soft target updates $\tau = 0.005$. Policy and magnitude updates subsample up to 2{,}048 transitions per batch.

\noindent\textbf{Evaluation Metrics.}
All metrics are computed in pixel space by rescaling normalized coordinates to the original $1264 \times 902$ resolution. We report Average Displacement Error (ADE), the mean Euclidean distance across all prediction steps; Final Displacement Error (FDE), the Euclidean distance at the last step; and discrete Fr\'{e}chet Distance (FD), measuring global trajectory shape similarity. Lower values indicate better performance for all three metrics.
\section{Results}
\label{sec:results}

\begin{table}[t]
\centering
\caption{Quantitative comparison on the test set under two setting (Obs = 6, Pred = 3 and Obs = 3, Pred = 6). Specifically, Obs = 6 denotes an observed surgical trajectory comprising 6 annotated keyframes, whereas Pred = 3 represents the prediction of the future trajectory containing 3 keyframes. All metrics are evaluated in pixel space (lower is better), with the best results highlighted in \textbf{bold} and second best in \textbf{underline}.}
\label{tab:qualitative_result}
\renewcommand{\arraystretch}{1.2} 
    \begin{tabular}{l ccc ccc} 
    \toprule
    \multirow{2}{*}[-0.5ex]{\textbf{Method}} 
      & \multicolumn{3}{c}{\textbf{Obs $=$ 6, Pred $=$ 3}}
      & \multicolumn{3}{c}{\textbf{Obs $=$ 3, Pred $=$ 6}} \\
      \cmidrule(lr){2-4} \cmidrule(lr){5-7} 
      & ADE ($\downarrow$) & FDE ($\downarrow$) & FD ($\downarrow$)
      & ADE ($\downarrow$) & FDE ($\downarrow$) & FD ($\downarrow$) \\
    \midrule
    BC~\cite{bain1999bc} & \underline{128.15}  & \underline{146.24} & \underline{156.83} & \underline{137.06}  & \underline{184.69} & \underline{195.13} \\
    GAIL~\cite{ho2016generative} & 269.79 & 282.99 & 305.86 & 249.19 & 254.41 & 318.82 \\
    IBC~\cite{florence2022implicit} & 243.97 & 262.12 & 285.60 & 225.74 & 260.49 & 296.82 \\
    iDiff-IL~\cite{li2023imitation} & 187.38 & 207.32 & 220.21 & 223.75 & 259.90 & 296.62 \\
    CondDiff~\cite{li2023imitation} & 165.20 & 189.47 & 200.31 & 184.15 & 236.84 & 255.62 \\
    MID~\cite{gu2022stochastic} & 151.23 & 165.24 & 192.79 & 172.67 & 229.83 & 246.65 \\
    \midrule
    \textbf{Ours} & \textbf{55.29} & \textbf{84.23} & \textbf{86.00} & \textbf{94.85} & \textbf{154.68} & \textbf{158.46} \\
    \bottomrule
    \end{tabular}
\end{table}

\subsection{Comparison with Baselines}
\label{subsec:comparison_baselines}

\begin{figure}
    \centering
    \includegraphics[width=0.90\linewidth]{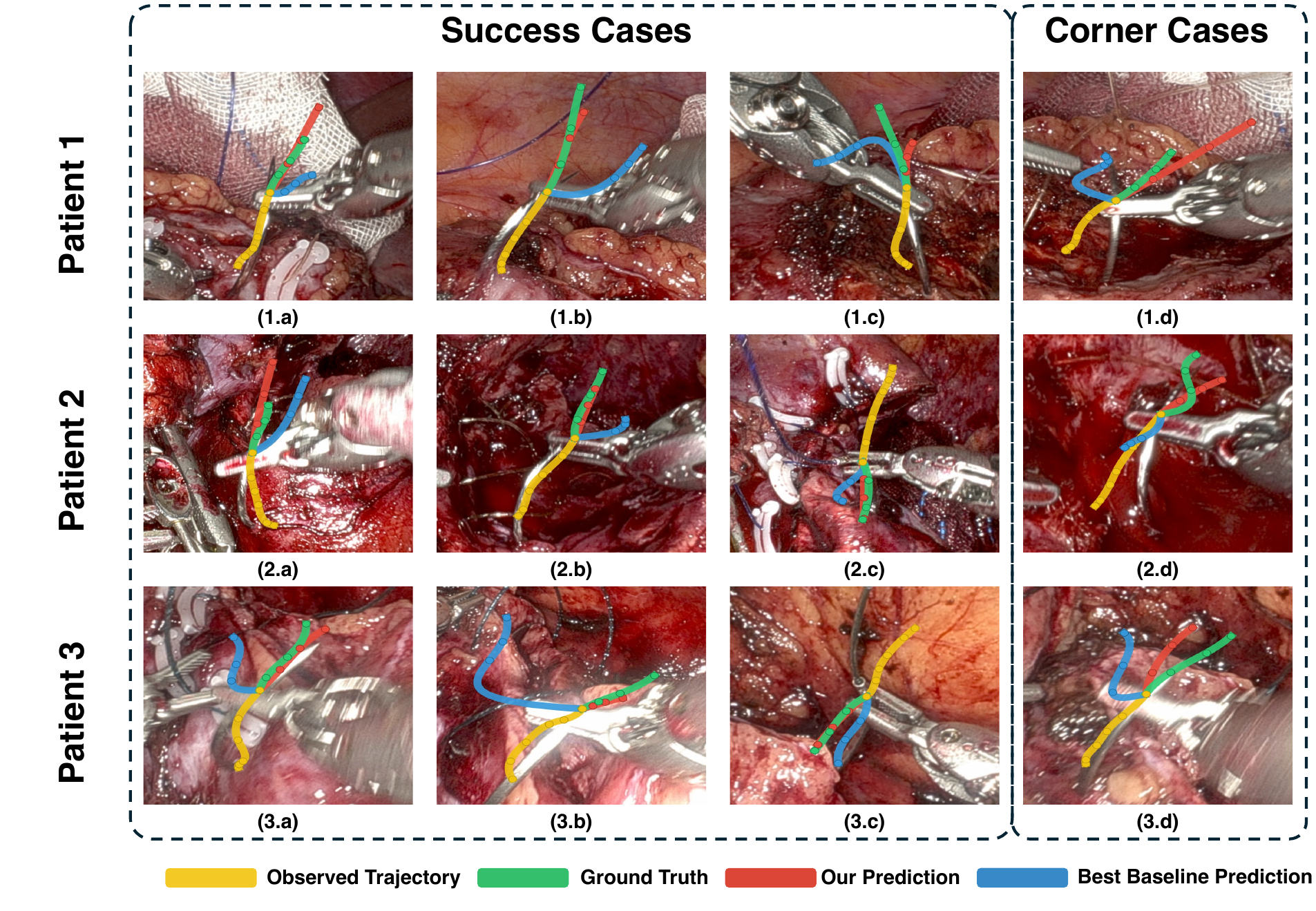}
    \caption{Qualitative comparison of predicted trajectories on the testset. The \textbf{yellow} curve denotes the observed trajectory, the \textbf{green} curve represents the ground truth future trajectory, the \textbf{red} curve shows the prediction from our \methodname\, and the \textbf{blue} curve indicates the best baseline prediction.}
    \label{fig:visualized_result}
\end{figure}

We compare \methodname\ against following baselines: (1)~Behavioral Cloning (BC)~\cite{bain1999bc}, which encodes stacked frames via a U-Net~\cite{ronneberger2015u} and regresses future coordinates with an MLP; (2)~Generative Adversarial Imitation Learning (GAIL)~\cite{ho2016generative}, which generates trajectories conditioned on a random latent vector with a discriminator distinguishing real from generated pairs; (3)~Implicit Behavioral Cloning (IBC)~\cite{florence2022implicit}, an energy based model trained with InfoNCE loss and sampled via Langevin MCMC at inference; (4)~Implicit Diffusion Policy (iDiff-IL)~\cite{li2023imitation}, a joint image trajectory diffusion method using a dual-head UNet to denoise both spaces simultaneously; (5)~Conditional Diffusion Policy (CondDiff)~\cite{li2023imitation}, a variant applying DDPM denoising in trajectory space only, conditioned on the image; and (6)~Motion Indeterminacy Diffusion (MID)~\cite{gu2022stochastic}, a latent-space diffusion method that encodes frames into a context vector and applies a Transformer-based DDPM denoiser for trajectory generation.

\textbf{Quantitative Results.}
Table~\ref{tab:qualitative_result} reports results under two experimental settings. \methodname\ consistently outperforms all baselines across all metrics. With Obs$=$6, Pred$=$3, \methodname\ achieves an ADE of 55.29 (56.8\% reduction over BC at 128.15) and FD of 84.23 (vs.\ 146.24 for BC), indicating substantially better trajectory shape fidelity. Under the more challenging Obs$=$3, Pred$=$6 setting, all methods degrade as historical observational information decreases, yet ours remains the best, achieving ADE 94.85, FDE 154.68 and FD 158.46. \methodname\ maintains clear advantages in ADE (30.7\% reduction over BC), FDE (16.2\% reduction) and FD (18.7\% reduction), demonstrating that the CQL-based policy produces more accurate trajectories with better shape consistency even under sparse observations. The goal-conditioned navigation formulation proves particularly beneficial when visual context is limited, as explicit target guidance compensates for reduced observational evidence. To further isolate the contribution of conservative offline reinforcement learning, we remove CQL from the training objective and observe 23.5\%, 18.6\%, and 18.1\% degradation in ADE, FDE, and FD, respectively. This indicates that the performance gain is not merely attributable to the architecture or preprocessing, but is substantially supported by conservative offline RL.

\textbf{Qualitative Results.}
Fig.~\ref{fig:visualized_result} illustrates representative predicted trajectories overlaid on surgical images, alongside the corresponding ground-truth paths and prediction results from selected state-of-art baseline method. As observed across diverse suturing scenarios from multiple patients, the baseline method frequently struggles to capture the complex spatial dynamics of the surgical instruments, resulting in predicted trajectories that significantly deviate from the true paths. In contrast, our proposed approach consistently maintains high fidelity to the ground truth. The proposed model generates smooth trajectories that conform to the general curvature of the suturing path.
\section{Conclusion}
\label{sec:conclusion}

This paper introduces \methodname, a novel framework that reformulates surgical trajectory prediction in pixel space based on goal-conditioned offline reinforcement learning using Conservative Q-Learning. By requiring only 9 keyframe annotations per trajectory and operating without robot kinematics, the framework demonstrates broad applicability to existing clinical video archives. Experimental results on 1,158 trajectories demonstrate that SutureFormer significantly outperforms best diffusion and imitation learning baselines, achieving up to a 56.8\% reduction in ADE. Future work will extend SutureFormer to broader laparoscopic procedures and validate its performance through ex-vivo porcine experiments on robotic platforms, facilitating its transition toward precise, advancing its translation toward real-world cognitive surgical assistance.
\subsubsection*{Acknowledgments.}
This work was supported by the University of Macau under Grants SRG2024-00056-FST, 0078/\allowbreak 2024/\allowbreak RIB2, and FST/\allowbreak SP01/\allowbreak 2024, the Dr.~Stanley Ho Medical Development Foundation under Grant SHMDF-AI/\allowbreak 2026/\allowbreak 002, the Natural Science Foundation of China under Grant 62373243, the Shanghai Municipal Health Commission Smart Healthcare Project under Grant 2025ZHYL021, and the Shanghai Tongren Hospital Medical-Engineering Collaboration Project under Grant lhyjzx2024-\allowbreak xm03. The surgical records of robot-assisted partial nephrectomy used in this work were collected by the Chinese PLA General Hospital and annotated by two expert surgeons.

\subsubsection*{Disclosure of Interests.}
The authors have no competing interests.

\bibliographystyle{splncs04}
\bibliography{references}

@INPROCEEDINGS{shi2022recognition,
    author={Shi, Chang and Zheng, Yi and Fey, Ann Majewicz},
    booktitle={IROS}, 
    title={Recognition and Prediction of Surgical Gestures and Trajectories Using Transformer Models in Robot-Assisted Surgery}, 
    year={2022},
    volume={},
    number={},
    pages={8017-8024},
    doi={10.1109/IROS47612.2022.9981611}}

@INPROCEEDINGS{qin2020davincinet,
    author={Qin, Yidan and Feyzabadi, Seyedshams and Allan, Max and Burdick, Joel W. and Azizian, Mahdi},
    booktitle={IROS}, 
    title={daVinciNet: Joint Prediction of Motion and Surgical State in Robot-Assisted Surgery}, 
    year={2020},
    volume={},
    number={},
    pages={2921-2928},
    doi={10.1109/IROS45743.2020.9340723}}

@INPROCEEDINGS{weerasinghe2024multimodal,
    author={Weerasinghe, Keshara and Reza Roodabeh, Seyed Hamid and Hutchinson, Kay and Alemzadeh, Homa},
    booktitle={ICRA}, 
    title={Multimodal Transformers for Real-Time Surgical Activity Prediction}, 
    year={2024},
    volume={},
    number={},
    pages={13323-13330},
    doi={10.1109/ICRA57147.2024.10611048}}

@inproceedings{kumar2020cql,
    author = {Kumar, Aviral and Zhou, Aurick and Tucker, George and Levine, Sergey},
    title = {Conservative {Q}-learning for offline reinforcement learning},
    year = {2020},
    isbn = {9781713829546},
    publisher = {Curran Associates Inc.},
    address = {Red Hook, NY, USA},
    booktitle = {NeurIPS},
    articleno = {100},
    numpages = {13},
    location = {Vancouver, BC, Canada},
    }

@misc{yang2017medical,
    title={Medical robotics—Regulatory, ethical, and legal considerations for increasing levels of autonomy},
    author={Yang, Guang-Zhong and Cambias, James and Cleary, Kevin and Daimler, Eric and Drake, James and Dupont, Pierre E and Hata, Nobuhiko and Kazanzides, Peter and Martel, Sylvain and Patel, Rajni V and others},
    journal={Science robotics},
    volume={2},
    number={4},
    pages={eaam8638},
    year={2017},
    publisher={American Association for the Advancement of Science}
}

@article{attanasio2021autonomy,
    author = "Attanasio, Aleks and Scaglioni, Bruno and De Momi, Elena and Fiorini, Paolo and Valdastri, Pietro",
    title = "Autonomy in Surgical Robotics", 
    journal= "Annual Review of Control, Robotics, and Autonomous Systems",
    year = "2021",
    volume = "4",
    number = "Volume 4, 2021",
    pages = "651-679",
    doi = "https://doi.org/10.1146/annurev-control-062420-090543",
    url = "https://www.annualreviews.org/content/journals/10.1146/annurev-control-062420-090543",
    publisher = "Annual Reviews",
    issn = "2573-5144",
    type = "Journal Article",
    }

@inproceedings{pomerleau1989alvinn,
    author = {Pomerleau, Dean A.},
    title = {{ALVINN}: an autonomous land vehicle in a neural network},
    year = {1988},
    publisher = {MIT Press},
    address = {Cambridge, MA, USA},
    booktitle = {NIPS},
    pages = {305–313},
    numpages = {9},
    series = {NIPS'88}
    }

@inproceedings{li2023imitation,
    author = {Li, Jianan and Jin, Yueming and Chen, Yueyao and Yip, Hon-Chi and Scheppach, Markus and Chiu, Philip Wai-Yan and Yam, Yeung and Meng, Helen Mei-Ling and Dou, Qi},
    title = {Imitation Learning from Expert Video Data for Dissection Trajectory Prediction in Endoscopic Surgical Procedure},
    year = {2023},
    isbn = {978-3-031-43995-7},
    publisher = {Springer-Verlag},
    address = {Berlin, Heidelberg},
    url = {https://doi.org/10.1007/978-3-031-43996-4_47},
    doi = {10.1007/978-3-031-43996-4_47},
    booktitle = {MICCAI},
    pages = {494–504},
    numpages = {11},
    location = {Vancouver, BC, Canada}
    }

@inproceedings{bain1999bc,
    author = {Bain, Michael and Sammut, Claude},
    title = {A Framework for Behavioural Cloning},
    year = {1999},
    isbn = {0198538677},
    publisher = {Oxford University},
    address = {GBR},
    booktitle = {Machine Intelligence 15, Intelligent Agents [St. Catherine's College, Oxford, July 1995]},
    pages = {103–129},
    numpages = {27}
    }

@inproceedings{ho2016generative,
    author = {Ho, Jonathan and Ermon, Stefano},
    title = {Generative adversarial imitation learning},
    year = {2016},
    isbn = {9781510838819},
    publisher = {Curran Associates Inc.},
    address = {Red Hook, NY, USA},
    booktitle = {NIPS},
    pages = {4572–4580},
    numpages = {9},
    location = {Barcelona, Spain},
    series = {NIPS'16}
    }

@InProceedings{florence2022implicit,
    title = 	 {Implicit Behavioral Cloning},
    author =       {Florence, Pete and Lynch, Corey and Zeng, Andy and Ramirez, Oscar A and Wahid, Ayzaan and Downs, Laura and Wong, Adrian and Lee, Johnny and Mordatch, Igor and Tompson, Jonathan},
    booktitle = 	 {Proceedings of the 5th Conference on Robot Learning},
    pages = 	 {158--168},
    year = 	 {2022},
    editor = 	 {Faust, Aleksandra and Hsu, David and Neumann, Gerhard},
    volume = 	 {164},
    series = 	 {PMLR},
    month = 	 {08--11 Nov},
    publisher =    {PMLR},
    url = 	 {https://proceedings.mlr.press/v164/florence22a.html},
}

@INPROCEEDINGS{gu2022stochastic,
    author={Gu, Tianpei and Chen, Guangyi and Li, Junlong and Lin, Chunze and Rao, Yongming and Zhou, Jie and Lu, Jiwen},
    booktitle={2022 IEEE/CVF Conference on Computer Vision and Pattern Recognition (CVPR)}, 
    title={Stochastic Trajectory Prediction via Motion Indeterminacy Diffusion}, 
    year={2022},
    volume={},
    number={},
    pages={17092-17101},
    doi={10.1109/CVPR52688.2022.01660}}

@article{maier2017surgical,
    title={Surgical data science for next-generation interventions},
    author={Maier-Hein, Lena and Vedula, Swaroop S and Speidel, Stefanie and Navab, Nassir and Kikinis, Ron and Park, Adrian and Eisenmann, Matthias and Feussner, Hubertus and Forestier, Germain and Giannarou, Stamatia and others},
    journal={Nature Biomedical Engineering},
    volume={1},
    number={9},
    pages={691--696},
    year={2017},
    publisher={Nature Publishing Group UK London}
    }

@article{jin2022trans,
    title={Trans-SVNet: hybrid embedding aggregation Transformer for surgical workflow analysis},
    author={Jin, Yueming and Long, Yonghao and Gao, Xiaojie and Stoyanov, Danail and Dou, Qi and Heng, Pheng-Ann},
    journal={International Journal of Computer Assisted Radiology and Surgery},
    volume={17},
    number={12},
    pages={2193--2202},
    year={2022},
    publisher={Springer}
}

@article{nwoye2022rendezvous,
    title={Rendezvous: Attention mechanisms for the recognition of surgical action triplets in endoscopic videos},
    author={Nwoye, Chinedu Innocent and Yu, Tong and Gonzalez, Cristians and Seeliger, Barbara and Mascagni, Pietro and Mutter, Didier and Marescaux, Jacques and Padoy, Nicolas},
    journal={Medical Image Analysis},
    volume={78},
    pages={102433},
    year={2022},
    publisher={Elsevier}
}

@article{cai2021vision,
    title={Vision-based autonomous car racing using deep imitative reinforcement learning},
    author={Cai, Peide and Wang, Hengli and Huang, Huaiyang and Liu, Yuxuan and Liu, Ming},
    journal={IEEE Robotics and Automation Letters},
    volume={6},
    number={4},
    pages={7262--7269},
    year={2021},
    publisher={IEEE}
}

@article{tamar2016value,
    title={Value iteration networks},
    author={Tamar, Aviv and Wu, Yi and Thomas, Garrett and Levine, Sergey and Abbeel, Pieter},
    journal={Advances in neural information processing systems},
    volume={29},
    year={2016}
}

@article{simon1972theories,
    title={Theories of bounded rationality},
    author={Simon, Herbert A and others},
    journal={Decision and organization},
    volume={1},
    number={1},
    pages={161--176},
    year={1972},
    publisher={Amsterdam}
}

@book{de1978practical,
    title={A practical guide to splines},
    author={De Boor, Carl and De Boor, Carl},
    volume={27},
    year={1978},
    publisher={springer New York}
}

@article{bojarski2016end,
  title={End to end learning for self-driving cars},
  author={Bojarski, Mariusz and Del Testa, Davide and Dworakowski, Daniel and Firner, Bernhard and Flepp, Beat and Goyal, Prasoon and Jackel, Lawrence D and Monfort, Mathew and Muller, Urs and Zhang, Jiakai and others},
  journal={arXiv preprint arXiv:1604.07316},
  year={2016}
}

@inproceedings{ronneberger2015u,
  title={U-net: Convolutional networks for biomedical image segmentation},
  author={Ronneberger, Olaf and Fischer, Philipp and Brox, Thomas},
  booktitle={MICCAI},
  pages={234--241},
  year={2015},
  organization={Springer}
}
\end{document}